\documentclass{article}

 \usepackage[main, final]{neurips_2026}

\usepackage[utf8]{inputenc} 
\usepackage[T1]{fontenc}    
\usepackage{hyperref}       
\usepackage{url}            
\usepackage{booktabs}       
\usepackage{amsfonts}       
\usepackage{nicefrac}       
\usepackage{microtype}      
\usepackage{xcolor}         
\usepackage{amsmath}
\usepackage{amssymb}
\usepackage{mathtools}
\usepackage{amsthm}
\usepackage{algorithm}
\usepackage{algpseudocode}
\usepackage{wrapfig}
\theoremstyle{plain}
\newtheorem{theorem}{Theorem}[section]
\newtheorem{proposition}[theorem]{Proposition}

\newtheorem{corollary}[theorem]{Corollary}
\theoremstyle{definition}
\newtheorem{definition}[theorem]{Definition}
\newtheorem{assumption}[theorem]{Assumption}
\theoremstyle{remark}

\title{Scale-Invariant Empirical-Bayes Laplace Approximation for ReLU Networks}

\author{%
  Shivam Pal\textsuperscript{1} and Piyush Rai\textsuperscript{1,2}\\
  \textsuperscript{1}Department of Computer Science and Engineering, IIT Kanpur\\
  \textsuperscript{2}Department of Intelligent Systems, IIT Kanpur\\
  \texttt{\{pshivam,piyush\}@cse.iitk.ac.in}
}

\begin{document}

\maketitle

\begin{abstract} 
ReLU networks are well-known to have continuous scale symmetries; infinitely many reparameterized weight vectors that represent the same function.\setcitestyle{authoryear, round}\citet{Dinh2017SharpMinima} have shown that, by reparameterizing, the geometry of the loss landscape (e.g., curvature) of these networks can be changed arbitrarily without affecting the underlying function. 
Thus, any quantity that depends on the network's geometry should be defined with care. In this paper, we show that the Empirical Bayes (EB) Laplace's approximation is not well-defined for ReLU networks, mainly due to Hessian dependence in computing the marginal likelihood, which is not invariant under ReLU scale symmetries. 
Thus, we can get different prior precisions, posterior covariances, and predictive uncertainties for a network having identical predictions under ReLU scaling. 
We define family of prior precision fields indexed by $\gamma$ and introduce \emph{$\gamma$-canonicalization}, which first converts the network to its \emph{canonical representation} by balancing the ratio of incoming to outgoing weights, and then fits the Empirical Bayes (EB) Laplace to the canonical representation. This procedure is equivalent to transporting the isotropic Gaussian prior from the canonical representation to the original coordinates, by producing a ReLU-scale-compatible precision field. This makes the marginal likelihood invariant under ReLU scaling, while allowing $\gamma$ to be selected by maximizing the marginal likelihood. The resulting EB Laplace approximation is well-defined on ReLU scale-equivalence classes and yields consistent prior precision selection and predictive uncertainties across functionally identical parameterizations.

\end{abstract}  
\section{Introduction}
\label{sec-introduction}
The Laplace's approximation is a simple post-hoc method for approximate Bayesian inference in deep neural networks. It fits a local Gaussian posterior whose mean is a maximum-a-posteriori (MAP) solution, and covariance is determined by 
curvature and prior precision~\citep{Mackay1992,ritter2018a,daxberger2021laplace}. The prior precision is often selected using marginal likelihood, which can be estimated solely from training data and without any validation data. This is also known as empirical Bayes (EB) or type-II maximum likelihood learning. 
Recent progress in structured curvature approximations, linearized prediction, and software support~\citep{ritter2018a,
daxberger2021laplace,ImmerBFRK21} has made the EB Laplace approximation an attractive method for post-hoc uncertainty estimation and Bayesian model selection.

In this paper, we study the transformation properties of marginal likelihood under ReLU scale symmetries. 
For ReLU networks, the marginal likelihood is not well-defined on the learned
predictor. Due to the positive homogeneity property of ReLU, multiplying the incoming and outgoing weights of a hidden unit by $\alpha>0$ and $1/\alpha$, respectively, makes the network function unchanged. These transformations generate ReLU \emph{scale
orbits}; continuous families of parameter vectors having the same
function~\citep{Neyshabur2015Norm, NeyshaburSS15, Dinh2017SharpMinima}. Along
such an orbit, the marginal likelihood is not invariant mainly due to the log-determinant of the curvature and the norm of the chosen orbit which do not transform compatibly. Consequently, standard EB Laplace can give different
marginal likelihoods,  prior precisions, and linearized predictive
uncertainties to the parameterizations that represent exactly the same
predictor.

We fix this problem by introducing \emph{$\gamma$-canonicalization} where $\gamma\in[0,1]$ controls the choice of canonical representation. Our procedure  first converts the network (MAP solution) to its canonical representation
by balancing the ratio of incoming and outgoing weights of the hidden units of the network, then
fitting the EB Laplace on the canonical representation. This generalizes the balanced and one-sided normalizations studied in
scale-symmetric networks~\citep{Stock2019EquiNorm,saul2023weightbalancing}.
This procedure is equivalent to transporting the isotropic Gaussian prior from the canonical representation to the original coordinates, by producing a ReLU-scale-compatible precision field. We show that, for every $\gamma$, the
resulting generalized marginal likelihood is invariant on the ReLU scale orbit. Thus, it makes the EB Laplace approximation explicitly depend on the learned function and not on arbitrary parametrization of the function that is obtained due to training artifacts. 
$\gamma$ is selected jointly with prior precision using generalized marginal likelihood.

Empirically, we further show that the standard EB Laplace varies substantially across functionally identical ReLU rescaling, while $\gamma$-canonicalized EB Laplace becomes invariant. We also find that $\gamma$-canonicalized EB Laplace gives better uncertainty estimates than standard Laplace, although our main guarantees are for the invariance. 
\section{Empirical-Bayes Laplace Approximation}
\label{sec:background}

Let $\mathbf{f}(\cdot\,;\mathbf{w})\colon\mathcal{X}\to\mathbb{R}^{C}$ be a
neural network with parameters $\mathbf{w}\in\mathbb{R}^{D}$, where $C$ is the
number of output classes and $D$ is the number of parameters. Let
$\mathcal{D}=\{(x_n,y_n)\}_{n=1}^{N}$ denote the training data. We assume that a
MAP solution $\hat{\mathbf{w}}$ is obtained by minimizing the regularized training loss, and  $\log p(\mathcal{D}\mid\mathbf{w})$ denotes
the log-likelihood evaluated at $\mathbf{w}$.

\textbf{Laplace Approximation.}
The Laplace approximation replaces the intractable posterior
$p(\mathbf{w}\mid\mathcal{D})$ by a Gaussian centered at $\hat{\mathbf{w}}$.
Given a likelihood-curvature approximation
$\mathbf{C}_{\hat{\mathbf{w}}}\in\mathbb{S}^{D}_{+}$ and a prior precision
$\lambda>0$, the approximate posterior is
$
    q(\mathbf{w})
    =
    \mathcal{N}\!\left(
        \mathbf{w}\mid
        \hat{\mathbf{w}},
        \boldsymbol{\Sigma}_{\hat{\mathbf{w}}}
    \right),
    \boldsymbol{\Sigma}_{\hat{\mathbf{w}}}
    =
    \left(
        \mathbf{C}_{\hat{\mathbf{w}}}+\lambda\mathbf{I}
    \right)^{-1}.
$
The exact Hessian is often not tractable, and also not guaranteed to be positive definite for neural networks. Common choices for curvature approximations are the generalized Gauss--Newton (GGN) matrix and the empirical
Fisher matrix~\citep{Mackay1992,ritter2018a,daxberger2021laplace}. The GGN and
Fisher are positive semi-definite by construction and are often preferred in
practice. We will assume the GGN approximation of curvature throughout the paper.

\textbf{Prior Precision and Empirical-Bayes Selection.}
The prior precision $\lambda$ controls the scale of the Gaussian posterior and affects the uncertainty estimates. It can be selected using a validation set. However a
common empirical-Bayes alternative is to select $\lambda$ by maximizing the approximate marginal likelihood on the training data. Up
to constants independent of $\lambda$, the log marginal
likelihood is
\begin{equation}
    \mathcal{L}_{\mathrm{std}}(\lambda;\hat{\mathbf{w}})
    =
    \log p(\mathcal{D}\mid\hat{\mathbf{w}})
    - \tfrac{1}{2}
      \log\left|
        \mathbf{C}_{\hat{\mathbf{w}}}+\lambda\mathbf{I}
      \right|
    + \tfrac{D}{2}\log\lambda
    - \tfrac{\lambda}{2}\|\hat{\mathbf{w}}\|^{2}.
    \label{eq:eb-objective}
\end{equation}
Note that for deep neural networks, the marginal likelihood cannot be computed exactly. Eq. ~\ref{eq:eb-objective} is only an approximation, widely used in Bayesian deep learning \citep{Mackay1992, ImmerBFRK21, Immer2021LocalLinearization, antoran2022linearised}.
The prior precision is then given by
\begin{equation}
    \lambda^{\star}(\hat{\mathbf{w}})
    \in
    \operatorname*{arg\,max}_{\lambda>0}
    \mathcal{L}_{\mathrm{std}}(\lambda;\hat{\mathbf{w}}).
    \label{eq:eb-precision}
\end{equation}
\textbf{Linearized Predictive Distribution.}
For a test input $x$, let
$\mathbf{J}_{\hat{\mathbf{w}}}(x)\in\mathbb{R}^{C\times D}$ denote the Jacobian
of $\mathbf{f}(\cdot\,;\mathbf{w})$ with respect to $\mathbf{w}$, evaluated at
$\hat{\mathbf{w}}$. The Jacobian-linearized distribution over network outputs is
Gaussian with mean 
$
    \boldsymbol{\mu}(x)=
    \mathbf{f}(x;\hat{\mathbf{w}}),
    $ and covariance $
    \mathbf{K}_{\lambda}(x)
    =
    \mathbf{J}_{\hat{\mathbf{w}}}(x)\,
    \boldsymbol{\Sigma}_{\hat{\mathbf{w}}}\,
    \mathbf{J}_{\hat{\mathbf{w}}}(x)^{\top}.
$

\section{Parameter-Space Symmetries in Neural Networks}
\label{sec:symmetries}

A parameter-space symmetry of a neural network is a transformation of the network
parameter that leaves the network function unchanged. Such symmetries
matter for post-hoc Bayesian inference because a quantity that changes under an exact
function-preserving transformation is a property of the parameter representative,
not of the learned function of the neural network.

\begin{definition}
A map $T\colon\mathbb{R}^{D}\to\mathbb{R}^{D}$ is a parameter-space symmetry
of $\mathbf{f}(\cdot\,;\mathbf{w})$ if
\begin{equation}
    \mathbf{f}(x;T(\mathbf{w}))
    =
    \mathbf{f}(x;\mathbf{w})
    \qquad
    \text{for all } x\in\mathcal{X}
    \text{ and } \mathbf{w}\in\mathbb{R}^{D}. \notag
\end{equation} 
\end{definition}

Given a set of symmetries $\mathcal{G}$, a scalar-valued function
$\phi(\mathbf{w})$ is \emph{invariant} under $\mathcal{G}$ if
$\phi(T(\mathbf{w}))=\phi(\mathbf{w})$ for all $T\in\mathcal{G}$. A
matrix-valued function $\mathbf{\Phi}(\mathbf{w})$ is \emph{equivariant} under
$\mathcal{G}$ if it transforms predictably under each $T\in\mathcal{G}$, meaning
there exists a known map $\rho_T$ such that
$\mathbf{\Phi}(T(\mathbf{w}))=\rho_T(\mathbf{\Phi}(\mathbf{w}))$. Invariance is
the special case where $\rho_T$ is the identity.

Neural networks have several exact parameter-space symmetries~\citep{zhao2026symmetry}. For example,
permutation symmetry, i.e., relabeling the hidden units of a layer and
applying the corresponding permutation to adjacent weight matrices leaves the
function unchanged~\citep{Hecht-Nielsen1990,entezari2022the,ainsworth2023git}.
In this work, we focus only on continuous symmetry specific to ReLU networks.

\subsection{ReLU Scale Orbits}
\label{sec:relu-orbits}

ReLU networks have continuous exact symmetries arising due to their homogeneity property. For any $\alpha>0$,
$
    \mathrm{ReLU}(\alpha z)
    =
    \alpha\,\mathrm{ReLU}(z).
$
Thus, the scale of a hidden unit can be redistributed between its incoming and
outgoing weights without changing the network function. Consider a feedforward
ReLU network with weight matrix
$\mathbf{W}^{(\ell)}\in\mathbb{R}^{d_{\ell}\times d_{\ell-1}}$ and bias vector
$\mathbf{b}^{(\ell)}\in\mathbb{R}^{d_{\ell}}$ at layer $\ell$. For a hidden unit
$(\ell,i)$ and a scalar $\alpha>0$, the rescaling
\begin{equation}
    \mathbf{W}^{(\ell)}[i,:]
    \leftarrow
    \alpha\,\mathbf{W}^{(\ell)}[i,:],
    \qquad
    \mathbf{b}^{(\ell)}[i]
    \leftarrow
    \alpha\,\mathbf{b}^{(\ell)}[i],
    \qquad
    \mathbf{W}^{(\ell+1)}[:,i]
    \leftarrow
    \alpha^{-1}\mathbf{W}^{(\ell+1)}[:,i]
    \label{eq:elementary-rescaling}
\end{equation}
leaves the network function unchanged.
Composing this over all hidden units creates a group $\mathcal{G}$ of
exact function-preserving transformations.
Let $\mathbf{w}\in\mathbb{R}^{D}$ be the flattened parameter and  $\boldsymbol{\alpha}=(\alpha_{\ell,i})>0$ collects one positive scalar per
hidden unit. Then, the corresponding transformation is $
    T_{\boldsymbol{\alpha}}(\mathbf{w})
    =
    \mathbf{D}_{\boldsymbol{\alpha}}\mathbf{w}.$
Here, $\mathbf{D}_{\boldsymbol{\alpha}}\in\mathbb{R}^{D\times D}$ is a diagonal matrix corresponding to $\boldsymbol{\alpha}$.
The \emph{ReLU scale orbit} of $\mathbf{w}$ is the equivalence class
$
    [\mathbf{w}]
    =
    \bigl\{
        T_{\boldsymbol{\alpha}}(\mathbf{w})
        \mid
        \boldsymbol{\alpha}>0
    \bigr\}.
$
Every element of $[\mathbf{w}]$ represents exactly the same function.

\subsection{Transformation Laws under ReLU Rescaling}
\label{sec:transformation-laws}

Let $\mathbf{v}=T_{\boldsymbol{\alpha}}(\mathbf{w})
=\mathbf{D}_{\boldsymbol{\alpha}}\mathbf{w}$ be a ReLU rescaling. By symmetry,
$
    \mathbf{f}(x;\mathbf{v})=\mathbf{f}(x;\mathbf{w}).
$
 Since $\boldsymbol{w} = \mathbf D_\alpha^{-1}\boldsymbol{v}$, and using the chain rule, the Jacobian transforms as follows.
\begin{equation}
\mathbf J_{\boldsymbol{v}}(x)
=\frac{\partial\mathbf f}{\partial\boldsymbol{v}}
=\frac{\partial\mathbf f}{\partial\boldsymbol{w}}\frac{\partial\boldsymbol{w}}{\partial\boldsymbol{v}}
=\mathbf J_{\boldsymbol{w}}(x)\,\mathbf D_\alpha^{-1}.
\label{eq:jacobian-transform}
\end{equation}
The generalized Gauss–Newton (GGN) curvature matrix can be written as $\mathbf C_{\boldsymbol{v}} = \sum_n \mathbf J_{\boldsymbol{v}}(x_n)^\top \boldsymbol\Lambda_n \mathbf J_{\boldsymbol{v}}(x_n)$. And this transforms
by congruence,
\begin{equation}
    \mathbf{C}_{\boldsymbol{v}}
    =
    \mathbf{D}_{\boldsymbol{\alpha}}^{-1}
    \mathbf{C}_{\boldsymbol{w}}
    \mathbf{D}_{\boldsymbol{\alpha}}^{-1}.
    \label{eq:curvature-transform}
\end{equation}
This congruence law also holds for standard KFAC curvature blocks under ReLU rescaling. We give the layerwise Kronecker-factor derivation in Appendix~\ref{proof:kfac}.
\section{Orbit Dependence of Standard Empirical-Bayes Laplace}
\label{sec:orbit-dependence}

\begin{definition}[Orbit consistency]
\label{def:orbit-consistency}
An empirical-Bayes Laplace procedure is \emph{orbit-consistent} if, for every
ReLU rescaling $\mathbf{v} = T_{\boldsymbol{\alpha}}(\mathbf{w})$,
the marginal likelihood, selected prior precision, and covariance of the predictive distribution are the same on the ReLU-scale orbit. 
\begin{equation}
    \mathcal{L}(\lambda;\mathbf{v}) = \mathcal{L}(\lambda;\mathbf{w}),
    \qquad
    \lambda^{\star}(\mathbf{v}) = \lambda^{\star}(\mathbf{w}),
    \qquad
    \mathbf{K}_{\lambda}(\mathbf{v};x) = \mathbf{K}_{\lambda}(\mathbf{w};x).
\end{equation}
\end{definition}

Standard EB Laplace approximation is not orbit-consistent. 
Let $\mathbf{v} = T_{\boldsymbol{\alpha}}(\mathbf{w}) =
\mathbf{D}_{\boldsymbol{\alpha}}\mathbf{w}$ be a ReLU rescaling. A sufficient condition for
$\mathcal{L}_{\mathrm{std}}$ to be orbit-consistent is that all four terms
in~\eqref{eq:eb-objective} are invariant under this rescaling. The
likelihood term satisfies this since the network function is unchanged,
$\log p(\mathcal{D}\mid\mathbf{v}) = \log p(\mathcal{D}\mid\mathbf{w})$.
However, the prior quadratic term satisfies
$\|\mathbf{v}\|^{2} = \|\mathbf{D}_{\boldsymbol{\alpha}}\mathbf{w}\|^{2} \neq
\|\mathbf{w}\|^{2}$ in general. And the
log-determinant term is also not invariant. Although the curvature transforms by
congruence~\eqref{eq:curvature-transform}, the isotropic component
$\lambda\mathbf{I}$ does not transform with it, so
$\mathbf{C}_{\mathbf{v}}+\lambda\mathbf{I}$ is not a congruence transform of
$\mathbf{C}_{\mathbf{w}}+\lambda\mathbf{I}$.
Thus, $\mathcal{L}_{\mathrm{std}}(\lambda;\mathbf{w})$ is not constant
on $[\mathbf{w}]$, and the prior precision $\lambda^{\star}(\mathbf{w})$ is selected by maximizing marginal likelihood. As we show below, the selected prior precision can also vary along the orbit. A detailed derivation of proposition \ref{prop:orbit-dependence} is provided in Appendix \ref{app:scalar-example}. 
\begin{proposition}[Orbit dependence of standard EB-Laplace]
\label{prop:orbit-dependence}
Consider the scalar ReLU regression model $f(x;\,a,c) = c\,\mathrm{ReLU}(ax)$
with $a,c>0$, Gaussian observation noise of variance $\sigma^2_y$, and inputs
$x_n>0$. The rescaling $(a,c)\mapsto(\alpha a,\,c/\alpha)$ for $\alpha>0$
preserves the realized predictor. Let $S=\sum_{n=1}^{N}x_n^2$,
$\kappa=S/\sigma^2_y$, and $q(\alpha)=(\alpha a)^2+(c/\alpha)^2$. The unique
positive stationary point of $\mathcal{L}_{\mathrm{std}}(\lambda;\alpha)$ in
$\lambda$ is
\begin{equation}
    \lambda^{\star}(\alpha)
    = \tfrac{-\kappa\,q(\alpha) + \sqrt{\kappa^2 q(\alpha)^2+4\kappa}}{2}.
    \label{eq:scalar-lambda}
\end{equation}
Since $q(\alpha)$ is non-constant on the scale orbit, $\lambda^{\star}(\alpha)$
varies with $\alpha$. Hence standard empirical-Bayes Laplace assigns different
marginal likelihoods, selected prior precisions, and linearized predictive
uncertainties to parameter vectors representing exactly the same function.
\end{proposition}

\section{Scale-Consistent Laplace}
\label{sec:scale-consistent-laplace}

The orbit-dependence of standard empirical-Bayes Laplace comes from a mismatch
between curvature and the isotropic prior. Under a ReLU rescaling
$\mathbf{v}=T_{\boldsymbol{\alpha}}(\mathbf{w})
=\mathbf{D}_{\boldsymbol{\alpha}}\mathbf{w}$, Jacobian-based curvature
transforms by congruence (\ref{eq:curvature-transform}),
whereas the standard prior precision $\lambda\mathbf{I}$ is fixed in the
original coordinates. In this section we identify the transformation law required
for prior precision to be compatible with ReLU scale orbits, and prove the
resulting invariance of marginal likelihood, selected prior precision and covariance of linearized prediction distribution. 

\subsection{Scale-Compatible Precision Fields}

A scalar precision $\lambda$ cannot make $\lambda\mathbf{I}$ transform like
curvature under arbitrary hidden-unit rescalings. We therefore allow the prior
precision to depend on the MAP solution.

\begin{definition}[Scale-compatible precision field]
\label{def:scale-compatible}
A map
$\mathbf{w}\mapsto\boldsymbol{\Lambda}(\mathbf{w})\in\mathbb{S}^{D}_{++}$
is \emph{scale-compatible} if, for every ReLU rescaling
$\mathbf{v}=T_{\boldsymbol{\alpha}}(\mathbf{w})
=\mathbf{D}_{\boldsymbol{\alpha}}\mathbf{w}$,
\begin{equation}
    \boldsymbol{\Lambda}(\mathbf{v})
    =
    \mathbf{D}_{\boldsymbol{\alpha}}^{-1}
    \boldsymbol{\Lambda}(\mathbf{w})
    \mathbf{D}_{\boldsymbol{\alpha}}^{-1}.
    \label{eq:scale-compatible}
\end{equation}
\end{definition}

Given such a field, define the generalized Laplace evidence
\begin{equation}
    \mathcal{L}_{\mathrm{gen}}(\mathbf{w})
    =
    \log p(\mathcal{D}\mid\mathbf{w})
    -
    \tfrac{1}{2}
    \log
    \left|
        \mathbf{C}_{\mathbf{w}}
        +
        \boldsymbol{\Lambda}(\mathbf{w})
    \right|
    +
    \tfrac{1}{2}
    \log
    \left|
        \boldsymbol{\Lambda}(\mathbf{w})
    \right|
    -
    \tfrac{1}{2}
    \mathbf{w}^{\top}
    \boldsymbol{\Lambda}(\mathbf{w})
    \mathbf{w}.
    \label{eq:gen-evidence}
\end{equation}
When $\boldsymbol{\Lambda}(\mathbf{w})=\lambda\mathbf{I}$, this reduces to the
standard isotropic objective in~\eqref{eq:eb-objective}.

\begin{theorem}[Evidence invariance]
\label{thm:evidence-invariance}
Let $\mathbf{v}=T_{\boldsymbol{\alpha}}(\mathbf{w})
=\mathbf{D}_{\boldsymbol{\alpha}}\mathbf{w}$ be a ReLU rescaling. Suppose that
the curvature satisfies~\eqref{eq:curvature-transform} and that
$\boldsymbol{\Lambda}$ is scale-compatible. Then
$
    \mathcal{L}_{\mathrm{gen}}(\mathbf{v})
    =
    \mathcal{L}_{\mathrm{gen}}(\mathbf{w}).
$
\end{theorem}
Proof of the theorem~\ref{thm:evidence-invariance} is given in Appendix~\ref{proof:evidence-invariance}.
If
$\{\boldsymbol{\Lambda}_{\eta}(\mathbf{w})\}_{\eta\in\mathcal{H}}$
is a family of scale-compatible precision fields, then
$\mathcal{L}_{\mathrm{gen}}(\mathbf{v};\eta)
=
\mathcal{L}_{\mathrm{gen}}(\mathbf{w};\eta)$
for every $\eta$. Therefore, the  set of hyperparameters that are obtained by maximizing the marginal likelihood will also be constant 
 on ReLU scale orbit. 
 
\subsection{Posterior Equivariance and Predictive Invariance}

Scale compatibility also gives the correct transformation rule for the posterior covariance and the covariance of the predictive distribution under the Laplace approximation. 
Let posterior covariance be
$
    \boldsymbol{\Sigma}_{\mathbf{w}}
    =
    \left(
        \mathbf{C}_{\mathbf{w}}
        +
        \boldsymbol{\Lambda}(\mathbf{w})
    \right)^{-1}$, and $
    \boldsymbol{\Sigma}_{\mathbf{v}}
    =
    \left(
        \mathbf{C}_{\mathbf{v}}
        +
        \boldsymbol{\Lambda}(\mathbf{v})
    \right)^{-1}.
$
Then
$
    \boldsymbol{\Sigma}_{\mathbf{v}}
    =
    \mathbf{D}_{\boldsymbol{\alpha}}
    \boldsymbol{\Sigma}_{\mathbf{w}}
    \mathbf{D}_{\boldsymbol{\alpha}}.
$
Thus, the posterior covariance is equivariant under scale-compatible precision field. This is the
expected behavior under transformation. The induced
Jacobian-linearized predictive covariance, however, is invariant.

\begin{theorem}[Linearized predictive invariance]
\label{thm:predictive-invariance}
Let $\mathbf{v}=T_{\boldsymbol{\alpha}}(\mathbf{w})
=\mathbf{D}_{\boldsymbol{\alpha}}\mathbf{w}$. Suppose that the curvature
satisfies~\eqref{eq:curvature-transform} and that
$\boldsymbol{\Lambda}$ is scale-compatible. Then, for every test input $x$,
\begin{equation}
    \mathbf{f}(x;\mathbf{v})
    =
    \mathbf{f}(x;\mathbf{w}),
    \qquad
    \mathbf{J}_{\mathbf{v}}(x)
    \boldsymbol{\Sigma}_{\mathbf{v}}
    \mathbf{J}_{\mathbf{v}}(x)^{\top}
    =
    \mathbf{J}_{\mathbf{w}}(x)
    \boldsymbol{\Sigma}_{\mathbf{w}}
    \mathbf{J}_{\mathbf{w}}(x)^{\top}.
\end{equation}
\end{theorem}

\begin{proof}
The mean is invariant because $\mathbf{v}$ and $\mathbf{w}$ represent the same
network function. For the covariance, using the transformation rule for Jacobian from Eq.~\ref{eq:jacobian-transform},
$\mathbf{J}_{\mathbf{v}}(x)
=
\mathbf{J}_{\mathbf{w}}(x)
\mathbf{D}_{\boldsymbol{\alpha}}^{-1}$
\begin{align}
    \mathbf{J}_{\mathbf{v}}(x)
    \boldsymbol{\Sigma}_{\mathbf{v}}
    \mathbf{J}_{\mathbf{v}}(x)^{\top}
    &=
    \mathbf{J}_{\mathbf{w}}(x)
    \mathbf{D}_{\boldsymbol{\alpha}}^{-1}
    \mathbf{D}_{\boldsymbol{\alpha}}
    \boldsymbol{\Sigma}_{\mathbf{w}}
    \mathbf{D}_{\boldsymbol{\alpha}}
    \mathbf{D}_{\boldsymbol{\alpha}}^{-1}
    \mathbf{J}_{\mathbf{w}}(x)^{\top} \notag \\
    &=
    \mathbf{J}_{\mathbf{w}}(x)
    \boldsymbol{\Sigma}_{\mathbf{w}}
    \mathbf{J}_{\mathbf{w}}(x)^{\top}. \notag
\end{align}
\end{proof}

\subsection{Transported Prior from a Canonical Section}
\label{sec:transported-prior}
We have now seen that, to make the EB Laplace scale-consistent, the precision field should be scale-compatible. Here we show how to construct such a precision field. Let's assume that
 $C_\gamma\colon\mathbb{R}^{D}\to\mathbb{R}^{D}$ selects a
canonical representative
$\bar{\mathbf{w}}=C_\gamma(\mathbf{w})$
from a ReLU scale orbit and satisfies the following property
\begin{equation}
    C_\gamma(T_{\boldsymbol{\alpha}}(\mathbf{w}))
    =
    C_\gamma(\mathbf{w})
    \
    \text{for every ReLU rescaling }T_{\boldsymbol{\alpha}}.
    \label{eq:section-property}
\end{equation}
The above property Eq.~\ref{eq:section-property} is also called a \emph{section} property and $C_\gamma$ is called \emph{canonical section}.
We assume that ReLU scale orbit is non-degenerate and the scale displacement between
$\mathbf{w}$ and $C_\gamma(\mathbf{w})$ is well defined. Since
$\mathbf{w}$ and $C_\gamma(\mathbf{w})$ lie on the same orbit, there exists a
positive diagonal matrix $\mathbf{D}(\mathbf{w};\gamma)$ such that
$
    \mathbf{w}
    =
    \mathbf{D}(\mathbf{w};\gamma)
    C_\gamma(\mathbf{w}).
    \label{eq:transport-map}
$
This matrix records the scale displacement from the canonical representative to
the original representative.
Using an isotropic Gaussian prior with precision $\lambda$ at the canonical
representative,
\begin{equation}
    p_\gamma(\bar{\mathbf{w}})
    \propto
    \exp\!\left(
        -\tfrac{\lambda}{2}
        \|\bar{\mathbf{w}}\|^2
    \right),
    \qquad
    \bar{\mathbf{w}}=C_\gamma(\mathbf{w}).
\end{equation}
Using
$\bar{\mathbf{w}}
=
\mathbf{D}(\mathbf{w};\gamma)^{-1}\mathbf{w}$,
the canonical quadratic penalty becomes
\begin{equation}
    \tfrac{\lambda}{2}
    \|\bar{\mathbf{w}}\|^2
    =
    \tfrac{1}{2}
    \mathbf{w}^{\top}
    \boldsymbol{\Lambda}_\gamma(\mathbf{w};\lambda)
    \mathbf{w},
    \quad\text{where} \quad
    \boldsymbol{\Lambda}_\gamma(\mathbf{w};\lambda)
    =
    \lambda\,
    \mathbf{D}(\mathbf{w};\gamma)^{-2}.
    \label{eq:transported-prior}
\end{equation}
Thus, quadratic penalty of an isotropic Gaussian prior in canonical coordinates is equivalent to that induced by a representative-dependent (transported), diagonal precision in the original coordinates.

\begin{proposition}[Scale compatibility of the transported prior]
\label{prop:scale-compatible}
Suppose $C_\gamma$ satisfies the section property~\eqref{eq:section-property}.
Then, for every $\lambda>0$, the transported precision field
$\boldsymbol{\Lambda}_\gamma(\mathbf{w};\lambda)$ defined
by~\eqref{eq:transported-prior} is scale-compatible.
\end{proposition}

\begin{corollary}[Orbit-consistent transported Laplace]
\label{cor:orbit-invariance}
Suppose $C_\gamma$ satisfies the section property~\eqref{eq:section-property}.
Then, for every fixed $\gamma$ and $\lambda>0$, the generalized evidence and
the Jacobian-linearized predictive distribution induced by
$\boldsymbol{\Lambda}_\gamma(\mathbf{w};\lambda)$ are invariant under ReLU
rescaling. Moreover, the evidence-maximizing set of prior precisions is the same
for all representatives of a ReLU scale orbit.
\end{corollary}

The proof of the Proposition~\ref{prop:scale-compatible} is given in Appendix~\ref{proof:compt_tp_prior}.
Combining Proposition~\ref{prop:scale-compatible} with
Theorems~\ref{thm:evidence-invariance} and~\ref{thm:predictive-invariance}
gives the section-level results.
The transported-prior framework reduces the remaining problem to constructing a
section $C_\gamma$ satisfying Eq.~\eqref{eq:section-property}. Next we give
such a construction via $\gamma$-canonicalization.

\section{\texorpdfstring{$\gamma$}{gamma}-Canonicalization}
\label{sec:gamma-canon}

 Any canonical section
$C_\gamma$ satisfying~\eqref{eq:section-property} makes
empirical-Bayes Laplace orbit-consistent.  Here, we introduce a practical algorithm for selecting a canonical section. The construction is based on a family of gauge conditions
that balance incoming and outgoing weight norms of the hidden units. First, we state the assumptions required for this construction and then give the iterative procedure for selecting the canonical representation.

\subsection{Canonical Gauge Condition}

A canonical section should not depend on the arbitrary representative returned
by MAP training. 
ReLU scale orbits provide a natural way to define such a section.
For each hidden unit, the scale symmetry redistributes the magnitude between the
incoming and outgoing weights. Fixing the relative
placement of this magnitude gives a gauge condition on the orbit. The parameter
$\gamma\in[0,1]$ controls where the canonical representative lies between
incoming-normalized, balanced, and outgoing-normalized representatives. This  construction is 
closely related to equi-normalization and weight-balancing methods for
scale-symmetric networks~\citep{Stock2019EquiNorm,saul2023weightbalancing}.
Without loss of generality we assume a feedforward ReLU network without any normalization layers.  Let us define the 
magnitude of incoming weight of hidden unit $(\ell,i)$ by 
\begin{equation}
    s^{(\ell)}_i(\mathbf{w})
    =
    \left(
        \bigl\|\mathbf{W}^{(\ell)}[i,:]\bigr\|^2
        +
        \bigl|b^{(\ell)}_i\bigr|^2
    \right)^{1/2}
    \label{eq:incoming-magnitude}
\end{equation}
and the magnitude of outgoing weight by
\begin{equation}
    t^{(\ell)}_i(\mathbf{w})
    =
    \bigl\|\mathbf{W}^{(\ell+1)}[:,i]\bigr\|.
    \label{eq:outgoing-magnitude}
\end{equation}
Under the elementary rescaling~\eqref{eq:elementary-rescaling} of the same unit
$(\ell,i)$ by $\alpha>0$, with all other unit scales held fixed, these magnitudes
transform as
$
    s^{(\ell)}_i \mapsto \alpha s^{(\ell)}_i,
    \
    t^{(\ell)}_i \mapsto \alpha^{-1} t^{(\ell)}_i.
$
Thus the elementary rescaling changes the ratio between incoming and outgoing
magnitudes while preserving their product. Across a
multilayer network, the gauge equations are coupled because rescaling the current
hidden units changes the adjacent hidden unit's incoming and outgoing weight magnitudes.
For $\gamma\in[0,1]$, the \emph{$\gamma$-canonical gauge condition} is given by
\begin{equation}
    \bigl(s^{(\ell)}_i(\bar{\mathbf{w}})\bigr)^{1-\gamma}
    =
    \bigl(t^{(\ell)}_i(\bar{\mathbf{w}})\bigr)^{\gamma}
    \qquad
    \text{for every hidden unit }(\ell,i).
    \label{eq:gauge-condition}
\end{equation}
For $\gamma=0$, the
condition becomes $s(\bar{\mathbf{w}})=1$, giving an
incoming-normalized representative. For $\gamma=1$, it becomes
$t(\bar{\mathbf{w}})=1$, giving an outgoing-normalized representative.
For $\gamma=0.5$, it becomes
$s(\bar{\mathbf{w}})=t(\bar{\mathbf{w}})$, giving a
balanced or minimum norm representative. 

\begin{assumption}[Canonical section]
\label{ass:section}
For each $\gamma\in[0,1]$ and each non-degenerate parameter vector,
there exists a deterministic representative
$\bar{\mathbf{w}}=C_\gamma(\mathbf{w})\in[\mathbf{w}]$ satisfying the
gauge condition~\eqref{eq:gauge-condition}. The resulting map
$C_\gamma$ is a section of the ReLU scale quotient
\begin{equation}
    C_\gamma(T_{\boldsymbol{\alpha}}(\mathbf{w}))
    =
    C_\gamma(\mathbf{w})
    \qquad
    \text{for every ReLU rescaling }T_{\boldsymbol{\alpha}}.
    \label{eq:gamma-section-property}
\end{equation}
\end{assumption}

Assumption~\ref{ass:section} requires the existence of a canonical representative.  In our implementation, we compute $C_\gamma$ by the iterative
algorithm~\ref{alg:canonicalize}.

\begin{algorithm}[t]
\caption{$\gamma$-Canonicalization}
\label{alg:canonicalize}
\begin{algorithmic}[1]
\Require parameters $\mathbf{w}$, gauge index $\gamma\in[0,1]$,
         tolerance $\varepsilon>0$
\Ensure representative $\bar{\mathbf{w}}=C_\gamma(\mathbf{w})$
\Repeat
    \For{each hidden unit $(\ell,i)$ in sweep order}
        \State compute $s^{(\ell)}_i(\mathbf{w})>0$ and $t^{(\ell)}_i(\mathbf{w})>0$ \Comment{Skip dead neuron}
        \State $\alpha \leftarrow
        \bigl(t^{(\ell)}_i(\mathbf{w})\bigr)^\gamma
        \bigl(s^{(\ell)}_i(\mathbf{w})\bigr)^{-(1-\gamma)}$
        \State $\mathbf{W}^{(\ell)}[i,:]\leftarrow
        \alpha\,\mathbf{W}^{(\ell)}[i,:]$
        \State $b^{(\ell)}_i\leftarrow \alpha\,b^{(\ell)}_i$
        \State $\mathbf{W}^{(\ell+1)}[:,i]\leftarrow
        \alpha^{-1}\mathbf{W}^{(\ell+1)}[:,i]$
    \EndFor
\Until{$\max_{\ell,i}|\log\alpha_{\ell,i}|<\varepsilon$}
\State \Return $\mathbf{w}$
\end{algorithmic}
\end{algorithm}

\subsection{Iterative Construction}

To compute the canonical representative, we repeatedly apply elementary
rescalings. Holding all other unit scales fixed, the rescaling factor that
enforces~\eqref{eq:gauge-condition} for unit $(\ell,i)$ is
\begin{equation}
    \alpha_{\ell,i}(\mathbf{w};\gamma)
    =
    \bigl(t^{(\ell)}_i(\mathbf{w})\bigr)^{\gamma}
    \bigl(s^{(\ell)}_i(\mathbf{w})\bigr)^{-(1-\gamma)}.
    \label{eq:update-rule}
\end{equation}
Applying~\eqref{eq:elementary-rescaling} with this factor sets the local
residual
$
    r_{\ell,i}(\mathbf{w};\gamma)
    =
    (1-\gamma)\log s^{(\ell)}_i(\mathbf{w})
    -
    \gamma\log t^{(\ell)}_i(\mathbf{w})
$
to zero for that unit. Since the residuals are coupled across layers for
interior values of $\gamma$, we sweep over hidden units and repeat until the
maximum update size is below a tolerance. 
We do not provide a proof of convergence for the $\gamma$-canonicalization algorithm. It is left for future work. For $\gamma=0.5$, the convergence proof is given by \citep{saul2023weightbalancing}. In the experiments, convergence is assessed by the
maximum gauge residual and by verifying that canonicalization preserves the
network outputs to numerical precision. We find the algorithm always converges within at most 200 sweeps. We use a first-to-last sweep order for $\gamma<1$ and a last-to-first sweep order for $\gamma=1$. The algorithm converges in a single sweep for $\gamma=0$ and $\gamma=1$.

\subsection{Selecting \texorpdfstring{$\gamma$}{gamma}}
\label{sec:selecting-gamma}

The parameter $\gamma$ specifies where the isotropic prior is placed on the
incoming and outgoing scale axis. It is therefore a prior-shape hyperparameter. Different values of
$\gamma$ correspond to different orbit-consistent priors.
Since $\gamma$ indexes a prior family, it can be selected by the same marginal
likelihood criterion used to select the prior precision. For a finite candidate
set $\Gamma\subseteq[0,1]$, we choose
\begin{equation}
    (\lambda^\star,\gamma^\star)
    \in
    \operatorname*{arg\,max}_{\lambda>0,\,\gamma\in\Gamma}
    \mathcal{L}_{\mathrm{gen}}(\mathbf{w};\lambda,\gamma).
    \label{eq:joint-selection}
\end{equation}
By Corollary~\ref{cor:orbit-invariance}, under
Assumption~\ref{ass:section}, the generalized evidence is invariant on
$[\mathbf{w}]$ for every fixed $(\lambda,\gamma)$. Hence the maximizing set of
$(\lambda,\gamma)$ values is the same for every representative of the ReLU scale
orbit. 
The choice $\gamma=0.5$ is geometrically natural and corresponds to the minimum norm canonical representation, However, it need not be
optimal under the marginal likelihood. The evidence can prefer a prior placed
closer to the incoming-normalized or outgoing-normalized representative. We
therefore treat $\gamma$ as a hyperparameter and select it from a grid search in the experiments.
\section{Experiments}

We evaluate $\gamma$-canonicalization on both small and large network architectures, including ResNets.  First, we perform an orbit-stress test on a Fashion-MNIST ReLU MLP to test whether the evidence, selected precision, and predictive uncertainty remain invariant across functionally identical ReLU rescaling. Next, we use a CIFAR-10 ReLU ConvNet with KFAC-Laplace to examine how $\gamma$-canonicalization affects predictive uncertainty in a convolution model. 
Although our theoretical guarantee is about invariance, in the next experiment we test whether $\gamma$-canonicalization can improve uncertainty estimates when the MAP representative is poorly scaled.

\paragraph{Orbit-stress test.}
We first test orbit consistency by applying post-hoc Laplace to different parameterizations of the same trained ReLU predictor. We train a two-hidden-layer ReLU MLP on Fashion-MNIST and generate $K=20$ function-preserving ReLU rescalings of the MAP solution by sampling unit-wise $\log\alpha_i\sim\mathcal{N}(0,1)$ and applying the elementary rescaling in Eq.~\ref{eq:elementary-rescaling}. All representatives preserve the original logits to numerical precision. We then fit full-network diagonal-GGN EB-Laplace to each representative, either in raw coordinates or after $\gamma$-canonicalization. Complete experimental details are given in Appendix~\ref{app:exp_details}.

\begin{table}[!htbp]
\caption{Orbit stress test on a single trained ReLU predictor. We draw \(K=20\)
function-preserving ReLU rescalings with \(\log\alpha_i \sim \mathcal{N}(0,1)\) and fit
diagonal-GGN EB-Laplace to each representative. Values are mean \(\pm\) standard deviation
across orbit samples.}
\label{tab:orbit_stress}
\centering
\scriptsize
\setlength{\tabcolsep}{2.5pt}
\begin{tabular}{lccccccc}
\toprule
Method
& \(\log\lambda^\star\)
& NLL
& ECE
& Acc.
& \(\log Z/10^4\)
& \(\|w\|^2\)
& \(\max\|\Delta p\|_\infty\) \\
\midrule
Standard EB
  & \(2.69{\pm}0.36\)
  & \(0.3906{\pm}0.0045\)
  & \(0.0682{\pm}0.0059\)
  & \(0.8718{\pm}0.0008\)
  & \(-8.80{\pm}0.89\)
  & \(2769{\pm}1016\)
  & \(2.62{\times}10^{-1}\) \\
\midrule
\(\gamma=0.00\)
  & \(5.07{\pm}0\)
  & \(0.3646{\pm}0\)
  & \(0.0387{\pm}0\)
  & \(0.8725{\pm}0\)
  & \(-4.72{\pm}0\)
  & \(156\)
  & \(8.35{\times}10^{-7}\) \\
\(\gamma=0.25\)
  & \(5.18{\pm}0\)
  & \(0.3644{\pm}0\)
  & \(0.0389{\pm}0\)
  & \(0.8726{\pm}0\)
  & \(-4.70{\pm}0\)
  & \(140\)
  & \(1.07{\times}10^{-6}\) \\
\(\gamma=0.50\)
  & \(5.29{\pm}0\)
  & \(0.3672{\pm}0\)
  & \(0.0428{\pm}0\)
  & \(0.8724{\pm}0\)
  & \(-4.95{\pm}0\)
  & \(134\)
  & \(8.35{\times}10^{-7}\) \\
\(\gamma=0.75\)
  & \(5.35{\pm}0\)
  & \(0.3723{\pm}0\)
  & \(0.0495{\pm}0\)
  & \(0.8721{\pm}0\)
  & \(-5.42{\pm}0\)
  & \(139\)
  & \(7.75{\times}10^{-7}\) \\
\(\gamma=1.00\)
  & \(5.32{\pm}0\)
  & \(0.3776{\pm}0\)
  & \(0.0562{\pm}0\)
  & \(0.8723{\pm}0\)
  & \(-5.94{\pm}0\)
  & \(156\)
  & \(7.75{\times}10^{-7}\) \\
\bottomrule
\end{tabular}
\end{table}

Tables~\ref{tab:orbit_stress} and~\ref{tab:orbit_summary} show that standard EB-Laplace is strongly orbit-dependent. Functionally identical representatives give different selected precisions, marginal likelihoods, and predictive probabilities. 
\begin{wraptable}[11]{r}{0.50\textwidth}
\vspace{-\intextsep}
\caption{Spread statistics of Orbit stress test. }
\label{tab:orbit_summary}
\centering
\scriptsize
\setlength{\tabcolsep}{4pt}
\begin{tabular}{lcc}
\toprule
Orbit statistic & Standard EB & \(\gamma\)-canonicalized EB \\
\midrule
\(\mathrm{std}(\log\lambda^\star)\)            & \(0.356\)              & \(0\) \\
\(\mathrm{range}(\log\lambda^\star)\)          & \(1.421\)              & \(0\) \\
Spread factor of \(\lambda^\star\)             & \(4.14{\times}\)       & \(1.00{\times}\) \\
\(\mathrm{range}(\log Z)\)                     & \(3.47{\times}10^4\) nats & \(<2{\times}10^{-2}\) nats \\
\(\mathrm{std}(\mathrm{NLL})\)                 & \(0.0045\)             & \(0\) \\
\(\mathrm{std}(\mathrm{ECE})\)                 & \(0.0059\)             & \(0\) \\
\(\max_{k,k'}\|p_k-p_{k'}\|_\infty\)           & \(2.62{\times}10^{-1}\) & \(\leq1.07{\times}10^{-6}\) \\
\(\max_{k}\|f_k-f_0\|_\infty\)                 & \(7.63{\times}10^{-6}\) & \(\leq9.54{\times}10^{-6}\) \\
\(\mathrm{Corr}(\log\lambda^\star,\|w\|^2)\)   & \(-0.971\)             & --- \\
\bottomrule
\end{tabular}
\end{wraptable}
In contrast, for every $\gamma$, canonicalized Laplace removes this variation to numerical precision. The evidence selects $\gamma=0.25$, while the balanced representative $\gamma=0.5$ gives the smallest parameter norm. However, it is not selected. It shows that generalized marginal likelihood is choosing a prior shape and not just minimizing the canonical parameter norm.
Random function-preserving rescalings of the same MAP solution substantially change standard EB-Laplace, confirming that explicit canonicalization is needed for orbit-consistency rather than relying on the scale representative returned by training.

\paragraph{CIFAR-10 ConvNet with KFAC-Laplace.}
We also evaluate $\gamma$-canonicalization with all-layer KFAC-Laplace on a
plain ReLU ConvNet trained on CIFAR-10. KFAC curvature satisfies the same congruence transformation law under ReLU rescaling, as shown in Appendix~\ref{proof:kfac}, so the scale-compatibility
argument applies beyond diagonal curvature. Table~\ref{tab:cifar10-convnet-kfac-main} shows the results. Over three independently trained
MAP seeds, standard KFAC-Laplace is strongly underconfident. Accuracy is
unchanged, but the mean predictive confidence is only $0.526$ for an accuracy of
$0.774$. Canonicalizing before fitting KFAC-Laplace, with $\gamma$ selected by
marginal likelihood within the canonicalized grid, selects $\gamma=0.5$ in all
three runs and substantially improves predictive quality. NLL decreases from
$0.877$ to $0.734$ and ECE from $0.248$ to $0.130$, with essentially unchanged
accuracy. Full experimental details and additional metrics are given in
Appendix~\ref{app:cifar10-kfac}.

\begin{table}[h]
\centering
\small
\caption{
CIFAR-10 ConvNet with all-layer KFAC-Laplace. Mean $\pm$ std over three MAP seeds.
}
\label{tab:cifar10-convnet-kfac-main}
\setlength{\tabcolsep}{6pt}
\begin{tabular}{lcccc}
\toprule
Method & Acc. & NLL & ECE & Conf. \\
\midrule
Standard KFAC-LA
    & $0.774 \pm 0.003$
    & $0.877 \pm 0.027$
    & $0.248 \pm 0.019$
    & $0.526 \pm 0.019$ \\
$\gamma$-canon KFAC-LA
    & $0.773 \pm 0.002$
    & $0.734 \pm 0.003$
    & $0.130 \pm 0.006$
    & $0.643 \pm 0.004$ \\
\bottomrule
\end{tabular}
\end{table}

\paragraph{CIFAR-100 ResNet with last-layer Laplace. }
ResNet networks do not have full ReLU-scale symmetry due to batch normalization layers. Thus, $\gamma$-canonicalization is not directly applicable. However, if we consider only the last-layer Laplace approximation, we can apply $\gamma$-canonicalization to get the canonical representation. Table~\ref{tab:resnet_orbit} shows the results for this. For these networks, the trained representative is already close to the canonical representative, so canonicalization does not improve the predictive performance. However, the marginal likelihood and prior precision still change substantially under rescaling, and canonicalization removes this variation to zero. 
While normalization layers help stabilize the scale in these networks during training, relying on them is not ideal. Relying on BatchNorm or specific weight decay values to accidentally balance the network is a fragile strategy. If a practitioner changes the optimizer, drops weight decay, or uses a novel architecture without standard normalization, standard EB-Laplace can fail silently, producing miscalibrated uncertainties. Canonicalization explicitly removes this fragility and guarantees that the Laplace fit is robust regardless of the training dynamics.
These results further demonstrate that canonicalization does not modify the predictions of an already well-balanced ResNet. Canonicalization yields behavior close to standard EB Laplace when the model is already well-scaled, and corrects the calibration issues in poorly scaled networks, as shown by the CIFAR-10 experiments \ref{tab:cifar10-convnet-kfac-main}.

\begin{table}[!htbp]
\caption{
Orbit-stress tests on CIFAR-100 with ResNet-18, ResNet-50, and ResNet-101
using last-layer EB-Laplace. For each MAP, we generate
$K=10$ function-preserving channel rescalings with independent
$\log\alpha_i\sim\mathcal{N}(0,1)$.
Values are mean $\pm$ standard deviation across orbit samples. 
}
\label{tab:resnet_orbit}
\setlength{\tabcolsep}{2.5pt}
\centering
\resizebox{\textwidth}{!}{%
\begin{tabular}{llrrrrrr}
\toprule
Model & Method
& $\log\lambda^\star$ & $\log Z/10^3$
& Acc. & NLL & ECE & Conf. \\
\midrule
ResNet-18 & MAP
& -- & --
& $0.7873$ & $0.8773$ & $0.0977$ & $0.8849$ \\
& Standard EB (orig.)
& $3.92$ & $-10.011$
& $0.7794$ & $0.7906$ & $0.0314$ & $0.7480$ \\
& Standard EB
& $2.89\pm0.13$ & $-27.060\pm1.307$
& $0.7720\pm0.0010$ & $1.1478\pm0.0487$
& $0.2711\pm0.0260$ & $0.5010\pm0.0252$ \\
\cmidrule(l){2-8}
& $\gamma=0.00$
& $2.54\pm0$ & $-10.087\pm0$
& $0.7793\pm0$ & $0.7918\pm0$
& $0.0327\pm0$ & $0.7466\pm0$ \\
& $\gamma=0.25$
& $3.04\pm0$ & $-10.049\pm0$
& $0.7794\pm0$ & $0.7912\pm0$
& $0.0321\pm0$ & $0.7473\pm0$ \\
& $\gamma=0.50$
& $3.54\pm0$ & $-10.032\pm0$
& $0.7794\pm0$ & $0.7909\pm0$
& $0.0318\pm0$ & $0.7476\pm0$ \\
& $\gamma=0.75$
& $4.04\pm0$ & $-10.034\pm0$
& $0.7794\pm0$ & $0.7910\pm0$
& $0.0318\pm0$ & $0.7476\pm0$ \\
& $\gamma=1.00$
& $4.53\pm0$ & $-10.055\pm0$
& $0.7793\pm0$ & $0.7912\pm0$
& $0.0319\pm0$ & $0.7474\pm0$ \\
\midrule
ResNet-50 & MAP
& -- & --
& $0.7917$ & $0.8801$ & $0.1088$ & $0.9004$ \\
& Standard EB (orig.)
& $4.18$ & $-12.421$
& $0.7870$ & $0.7592$ & $0.0341$ & $0.8128$ \\
& Standard EB
& $3.33\pm0.08$ & $-37.230\pm1.199$
& $0.7799\pm0.0005$ & $1.0358\pm0.0208$
& $0.2169\pm0.0128$ & $0.5631\pm0.0126$ \\
\cmidrule(l){2-8}
& $\gamma=0.00$
& $4.22\pm0$ & $-12.715\pm0$
& $0.7871\pm0$ & $0.7598\pm0$
& $0.0318\pm0$ & $0.8096\pm0$ \\
& $\gamma=0.25$
& $3.97\pm0$ & $-12.507\pm0$
& $0.7870\pm0$ & $0.7594\pm0$
& $0.0336\pm0$ & $0.8117\pm0$ \\
& $\gamma=0.50$
& $3.71\pm0$ & $-12.432\pm0$
& $0.7870\pm0$ & $0.7593\pm0$
& $0.0338\pm0$ & $0.8125\pm0$ \\
& $\gamma=0.75$
& $3.45\pm0$ & $-12.485\pm0$
& $0.7871\pm0$ & $0.7595\pm0$
& $0.0338\pm0$ & $0.8121\pm0$ \\
& $\gamma=1.00$
& $3.19\pm0$ & $-12.668\pm0$
& $0.7871\pm0$ & $0.7599\pm0$
& $0.0325\pm0$ & $0.8103\pm0$ \\
\midrule
ResNet-101 & MAP
& -- & --
& $0.8105$ & $0.8285$ & $0.1018$ & $0.9122$ \\
& Standard EB (orig.)
& $4.17$ & $-11.020$
& $0.8082$ & $0.7065$ & $0.0195$ & $0.8152$ \\
& Standard EB
& $3.34\pm0.09$ & $-33.887\pm1.141$
& $0.8014\pm0.0006$ & $1.0727\pm0.0245$
& $0.2753\pm0.0135$ & $0.5261\pm0.0135$ \\
\cmidrule(l){2-8}
& $\gamma=0.00$
& $4.31\pm0$ & $-11.279\pm0$
& $0.8083\pm0$ & $0.7076\pm0$
& $0.0187\pm0$ & $0.8119\pm0$ \\
& $\gamma=0.25$
& $3.98\pm0$ & $-11.075\pm0$
& $0.8082\pm0$ & $0.7067\pm0$
& $0.0192\pm0$ & $0.8145\pm0$ \\
& $\gamma=0.50$
& $3.65\pm0$ & $-11.019\pm0$
& $0.8083\pm0$ & $0.7065\pm0$
& $0.0196\pm0$ & $0.8151\pm0$ \\
& $\gamma=0.75$
& $3.31\pm0$ & $-11.105\pm0$
& $0.8081\pm0$ & $0.7071\pm0$
& $0.0188\pm0$ & $0.8140\pm0$ \\
& $\gamma=1.00$
& $2.96\pm0$ & $-11.342\pm0$
& $0.8079\pm0$ & $0.7085\pm0$
& $0.0177\pm0$ & $0.8110\pm0$ \\
\bottomrule
\end{tabular}%
}
\end{table}

\paragraph{Effect on model selection.} We compare two MLPs trained on CIFAR-10, with widths $64$ and $128$.
Marginal likelihood computed using canonical representation favors the width-$64$ model by
$12.4\times10^3$ nats.
For each ReLU rescaling strength $\sigma\in\{0.25,0.5,1.0\}$ (scale $\log\alpha \sim \mathcal{N}(0, \sigma^2)$), we create
$20$ versions of each model without changing its network predictions and
refit standard EB-Laplace, and compare all $400$ pairs.
As Table~\ref{tab:model_selection_orbits} shows, the mean evidence gap
changes from $11.1\times10^3$ to $53.7\times10^3$ nats.
Using the same rescaling strength for both models rarely changes their
ranking, because the wider model loses more evidence in this example.
However, when the width-$128$ model is kept at $\sigma=0.25$,
increasing the rescaling strength of the width-$64$ model reverses
the canonical ranking in $19.8\%$ of comparisons at $\sigma=0.5$
and $100\%$ at $\sigma=1.0$.
Thus, standard EB can change its model selection even though neither
network's predictions change. Canonicalization keeps the evidence gap
and model ranking fixed across these rescalings.
\begin{table}[t]
\centering
\scriptsize
\caption{
Standard EB model selection on CIFAR-10.
Evidence values are mean $\pm$ standard deviation over $20$
rescalings. Selection rates are computed over all $400$ pairs.
}
\label{tab:model_selection_orbits}
\begin{tabular}{crrrrr}
\toprule
& \multicolumn{3}{c}{Standard EB evidence ($10^3$ nats)}
& \multicolumn{2}{c}{Width $128$ selected (\%)} \\
\cmidrule(lr){2-4}
\cmidrule(lr){5-6}
$\sigma$
& Width $64$ & Width $128$ & Gap
& Both at $\sigma$ & Width $128$ at $0.25$ \\
\midrule
$0.25$
& $-110.6\pm0.8$ & $-121.7\pm0.7$
& $11.1$ & $0.0$ & $0.0$ \\
$0.50$
& $-119.3\pm2.9$ & $-136.4\pm2.7$
& $17.1$ & $0.0$ & $19.8$ \\
$1.00$
& $-160.6\pm11.9$ & $-214.3\pm16.9$
& $53.7$ & $0.2$ & $100.0$ \\
\bottomrule
\end{tabular}%
\end{table}

\section{Related Work}
\label{sec-related-work}

\paragraph{Laplace approximation and empirical-Bayes prior selection.}
The Laplace approximation is a classical tool for Bayesian model comparison,
with early evidence-based treatments established in the literature \citep{Mackay1992}.
Subsequent work makes it practical in neural networks through structured 
curvature approximations and scalable software \citep{ritter2018a, daxberger2021laplace}. 
Empirical-Bayes selection of prior precision by marginal-likelihood
maximisation has also been previously developed \citep{ImmerBFRK21} and extended to linearized
settings \citep{antoran2022linearised}. Our work does not propose a new
curvature approximation or scalable implementation. It identifies a structural
defect in the standard empirical-Bayes objective. It depends on the arbitrary parameter representative. 
We construct a corrected version that
is well-defined on ReLU scale-equivalence classes.

\paragraph{Classical Laplace theory and Singular models.}
Standard Laplace and Bayesian information criterion (BIC) asymptotics assume local identifiability and nonsingular quadratic expansion around an isolated mode \citep{Tierney1986, KassTierneyKadane1990}. ReLU networks violate these likelihood-level assumptions because of scale symmetry. Moreover, for deep neural networks, the Hessian need not be positive definite.  A Gaussian prior can make the posterior curvature positive definite, allowing a finite local Laplace approximation to be computed. However, the usual regular-model asymptotic interpretation of such approximation is no longer valid. \cite{mackay1998choice} has shown that the accuracy of a Laplace approximation can depend on the chosen parameter basis. For non-identifiable models, the regular BIC expansion is invalid. The asymptotic expansion of the log marginal likelihood instead involves the log canonical threshold and its multiplicity, which replace the regular complexity term and may introduce an additional contribution ~\citep{Watanabe2009}. Criteria such as wBIC \citep{watanabe2013} and sBIC \citep{Drton2017} are designed to account for this singular asymptotic behavior. We study a separate finite-sample defect of the empirical-Bayes Laplace surrogate used in practice. Our method does not correct the full singular asymptotics of neural-network marginal likelihoods. Rather, it makes that surrogate well-defined on ReLU scale orbit.

\paragraph{Scale symmetries and canonical representatives.}
Positive homogeneity of ReLU activations generates continuous families of
parameter vectors representing the same function. These symmetries have been 
studied extensively in the context of norm-based capacity control and 
path-normalized optimization \citep{Neyshabur2015Norm, NeyshaburSS15}.
Furthermore, it has been shown that ReLU reparameterizations change
sharpness measures without changing the represented function \citep{Dinh2017SharpMinima}, establishing
that coordinate-dependent quantities are not intrinsic properties of the
learned predictor. Our paper applies the same principle to empirical-Bayes
Laplace. The evidence and selected prior precision should not depend on an
arbitrary representative. To construct a canonical representative, we build
on iterative balancing procedures that minimize the weight norm over
a scale orbit \citep{Stock2019EquiNorm, saul2023weightbalancing}. Our $\gamma$-canonicalization generalizes these procedures. Rather than using balancing to improve optimization, our method uses it to define where an isotropic Gaussian prior is placed on each orbit, turning balancing
into a mechanism for scale-consistent empirical-Bayes inference.

\paragraph{Reparameterization invariance in approximate Bayesian inference.}
The most directly related work studies reparameterization invariance in approximate Bayesian inference,
showing that linearized Laplace behaves differently from sampled Laplace in
directions that do not change the function \citep{Roy2024ReparamInvariance}. 
A related argument suggests that Hessian-based quantities become inconsistent
under reparameterization when the relevant metric structure is ignored \citep{Kristiadi2023Geometry}.
This issue has been addressed by proposing a Riemannian
Laplace approximation that adapts the posterior geometry to the local shape
of the log-posterior \citep{Bergamin2023RiemannianLaplace}. Our contribution is narrower and more specific than
these geometric approaches. We focus exclusively on the exact continuous
scale symmetry of ReLU networks and derive the precise transformation law
the empirical-Bayes prior precision must satisfy for both the evidence and
the linearized predictive distribution to be invariant on scale orbits. \cite{antoranG-prior} study Laplace inference in networks with
normalisation layers and show that BatchNorm and related layers break the
posterior-volume interpretation of the standard evidence, proposing a
neural $g$-prior to address heterogeneous Jacobian scales. Our work
addresses a different invariance failure in plain ReLU networks, where
the empirical-Bayes evidence and selected prior precision depend on the
arbitrary representative of a ReLU scale orbit.

\paragraph{Weight-space symmetries in Bayesian neural networks.}
Discrete weight-space symmetries, such as hidden-unit permutations, have
received attention in Bayesian neural networks.
Previous studies analyze how permutation symmetries affect posteriors and propose constructions
that respect permutation invariance \citep{Rossi2023Permutation, Gelberg2024PermutationInvariant}. 
Other works explore both permutation and scale symmetries in weight-space posteriors \citep{Laurent2024SymmetryAwareBNN}.
Our work is complementary. We address the continuous ReLU scale symmetry
rather than discrete permutation symmetry, and focus specifically on the
empirical-Bayes evidence objective and prior precision selection rather than on the posterior geometry itself.
\section{Limitations and future work}
\label{sec:limit}
$\gamma$-canonicalization is only applicable to neural networks having activation functions with positive homogeneity property such as ReLU, leaky ReLU, PReLU, etc. Moreover, normalization layers such as batch normalization or layer normalization break the scale symmetry. Thus, $\gamma$-canonicalization will not work on the network having normalization layers. Extending our method to work on normalized networks would require deriving the appropriate symmetry group, canonical section, and transported prior for that architecture. We apply $\gamma$-canonicalization post-hoc after the network has been trained. It would be interesting to see how training dynamics change if we apply $\gamma$-canonicalization during training. We leave such investigations to future work. 

\section{Conclusion}
\label{sec-conclusion}

Standard EB Laplace approximation for ReLU networks is not invariant under ReLU scale symmetry. It gives different marginal likelihoods, prior precisions, and predictive distributions for function-preserving ReLU rescalings. We introduce $\gamma$-canonicalized EB Laplace approximation that fixes the ReLU scale dependence issue of standard EB Laplace. It selects a canonical representation and places an isotropic Gaussian prior on that, which is equivalent to transporting the isotropic Gaussian prior from the canonical representation to the original coordinates by producing a precision field.  This precision
field transforms compatibly with curvature, which makes the generalized marginal likelihood, prior precision, and predictive distribution invariant under ReLU rescaling. Moreover, Invariance of the marginal likelihood makes model selection via EB Laplace well-defined. Although our guarantee is for invariance, if the MAP solution has poor scale, $\gamma$-canonicalization can also give improved uncertainty estimates, as shown in our experiments.

\begin{ack}
SP acknowledges support from Visvesvaraya PhD fellowship. PR acknowledges support from Google Research and Satya \& Rao Remala Foundation.
\end{ack}

\bibliographystyle{unsrtnat} 
\bibliography{paper}
\appendix

\appendix
\newpage
\section{Proof of Proposition~\ref{prop:orbit-dependence}(Scalar Orbit Dependence)}
\label{app:scalar-example}

\paragraph{Model.}
Consider $f(x;\,a,c)=c\,\mathrm{ReLU}(ax)$ with $a,c>0$, Gaussian observation
noise of variance $\sigma^2_y$, and inputs $x_n>0$. Since $a>0$,
$\mathrm{ReLU}(ax_n)=ax_n$, so $f(x_n;a,c)=acx_n$. The rescaling
$(a,c)\mapsto(\alpha a,\,c/\alpha)$ preserves the realized predictor on the
dataset for every $\alpha>0$.

\paragraph{Curvature.}
The Jacobian at $(a,c)$ for a single input $x_n$ is
$\mathbf{J}_n(a,c)=(cx_n,\,ax_n)$.
For Gaussian regression with noise variance $\sigma^2_y$, the GGN matrix is
\begin{equation}
    \mathbf{G}(a,c)
    = \sum_{n=1}^{N}\frac{1}{\sigma^2_y}\mathbf{J}_n^{\top}\mathbf{J}_n
    = \frac{S}{\sigma^2_y}
      \begin{pmatrix}c^2 & ac \\ ac & a^2\end{pmatrix},
    \qquad S=\sum_{n=1}^{N}x_n^2.
\end{equation}
This matrix has rank one with nonzero eigenvalue
$\mu(\alpha)=\kappa q(\alpha)$, where $\kappa=S/\sigma^2_y$ and
$q(\alpha)=(\alpha a)^2+(c/\alpha)^2$.

\paragraph{Evidence along the orbit.}
Along the scale orbit, the likelihood is constant and
$\|(a(\alpha),c(\alpha))\|^2=q(\alpha)$.
The eigenvalues of $\mathbf{G}(\alpha)+\lambda\mathbf{I}$ are $\lambda$ and
$\mu(\alpha)+\lambda$, so
$|\mathbf{G}(\alpha)+\lambda\mathbf{I}|=\lambda(\mu(\alpha)+\lambda)$.
Up to constants independent of $\lambda$ and $\alpha$, the standard
empirical-Bayes objective reduces to
\begin{equation}
    \mathcal{L}_{\mathrm{std}}(\lambda;\alpha)
    = \frac{1}{2}\log\frac{\lambda}{\kappa q(\alpha)+\lambda}
      - \frac{\lambda}{2}q(\alpha).
    \label{eq:scalar-evidence}
\end{equation}

\paragraph{Empirical-Bayes optimum.}
Differentiating~\eqref{eq:scalar-evidence} with respect to $\lambda$ and
setting the derivative to zero gives
\begin{equation}
    \frac{1}{\lambda} - \frac{1}{\kappa q(\alpha)+\lambda} = q(\alpha).
\end{equation}
Combining fractions,
\begin{equation}
    \frac{\kappa }{\lambda(\kappa q(\alpha)+\lambda)} =1.
\end{equation}
Since $q(\alpha)>0$, this simplifies to
$\lambda(\kappa q(\alpha)+\lambda)=\kappa$,
or equivalently
\begin{equation}
    \lambda^2 + \kappa q(\alpha)\lambda - \kappa = 0.
\end{equation}
The unique positive root is
\begin{equation}
    \lambda^{\star}(\alpha)
    = \frac{-\kappa q(\alpha) + \sqrt{\kappa^2 q(\alpha)^2 + 4\kappa}}{2}.
\end{equation}

\paragraph{Non-invariance.}
The function $q(\alpha)=(\alpha a)^2+(c/\alpha)^2$ is strictly convex in
$\log\alpha$, minimized at $\alpha^2=c/a$, and unbounded as
$\alpha\to 0$ or $\alpha\to\infty$. Since $\lambda^{\star}$ is a strictly
decreasing function of $q$ (its derivative with respect to $q$ is negative
for all $q>0$), $\lambda^{\star}(\alpha)$ is non-constant along the orbit.
Hence standard empirical-Bayes Laplace assigns different marginal likelihoods,
selected prior precisions, and linearized predictive uncertainties to
parameter vectors representing exactly the same predictor. \qed

\section{Proof of theorem~\ref{thm:evidence-invariance} (Evidence invariance)}
\label{proof:evidence-invariance}
\begin{proof}
The likelihood term is invariant because $\mathbf{v}$ and $\mathbf{w}$
represent the same predictor. For the posterior precision,
\[
    \mathbf{C}_{\mathbf{v}}+\boldsymbol{\Lambda}(\mathbf{v})
    =
    \mathbf{D}_{\boldsymbol{\alpha}}^{-1}
    \bigl(
        \mathbf{C}_{\mathbf{w}}
        +
        \boldsymbol{\Lambda}(\mathbf{w})
    \bigr)
    \mathbf{D}_{\boldsymbol{\alpha}}^{-1}.
\]
Hence
\[
    \left|
        \mathbf{C}_{\mathbf{v}}+\boldsymbol{\Lambda}(\mathbf{v})
    \right|
    =
    |\mathbf{D}_{\boldsymbol{\alpha}}|^{-2}
    \left|
        \mathbf{C}_{\mathbf{w}}+\boldsymbol{\Lambda}(\mathbf{w})
    \right|.
\]
Similarly,
\[
    |\boldsymbol{\Lambda}(\mathbf{v})|
    =
    |\mathbf{D}_{\boldsymbol{\alpha}}|^{-2}
    |\boldsymbol{\Lambda}(\mathbf{w})|.
\]
The two log-determinant changes therefore cancel in
\eqref{eq:gen-evidence}. Finally,
\begin{align}
    \mathbf{v}^{\top}
    \boldsymbol{\Lambda}(\mathbf{v})
    \mathbf{v}
    &=
    (\mathbf{D}_{\boldsymbol{\alpha}}\mathbf{w})^{\top}
    \mathbf{D}_{\boldsymbol{\alpha}}^{-1}
    \boldsymbol{\Lambda}(\mathbf{w})
    \mathbf{D}_{\boldsymbol{\alpha}}^{-1}
    (\mathbf{D}_{\boldsymbol{\alpha}}\mathbf{w}) \notag \\
    &=
    \mathbf{w}^{\top}
    \boldsymbol{\Lambda}(\mathbf{w})
    \mathbf{w}. \notag
\end{align}\end{proof}

\section{Proof of Proposition~\ref{prop:scale-compatible} (Scale Compatibility of the Transported Prior) }
\label{proof:compt_tp_prior}
\begin{proof}
Let
$\mathbf{v}=T_{\boldsymbol{\alpha}}(\mathbf{w})
=\mathbf{D}_{\boldsymbol{\alpha}}\mathbf{w}$.
By the section property,
$C_\gamma(\mathbf{v})=C_\gamma(\mathbf{w})$. Using the construction in Section
\eqref{eq:transport-map},
\[
    \mathbf{v}
    =
    \mathbf{D}_{\boldsymbol{\alpha}}\mathbf{w}
    =
    \mathbf{D}_{\boldsymbol{\alpha}}
    \mathbf{D}(\mathbf{w};\gamma)
    C_\gamma(\mathbf{v}).
\]
Therefore
\[
    \mathbf{D}(\mathbf{v};\gamma)
    =
    \mathbf{D}_{\boldsymbol{\alpha}}
    \mathbf{D}(\mathbf{w};\gamma).
\]
Since these matrices are diagonal,
\begin{align}
    \boldsymbol{\Lambda}_\gamma(\mathbf{v};\lambda)
    &=
    \lambda
    \mathbf{D}(\mathbf{v};\gamma)^{-2} \notag \\
    &=
    \lambda
    \bigl(
        \mathbf{D}_{\boldsymbol{\alpha}}
        \mathbf{D}(\mathbf{w};\gamma)
    \bigr)^{-2} \notag \\
    &=
    \mathbf{D}_{\boldsymbol{\alpha}}^{-1}
    \bigl(
        \lambda
        \mathbf{D}(\mathbf{w};\gamma)^{-2}
    \bigr)
    \mathbf{D}_{\boldsymbol{\alpha}}^{-1} \notag \\
    &=
    \mathbf{D}_{\boldsymbol{\alpha}}^{-1}
    \boldsymbol{\Lambda}_\gamma(\mathbf{w};\lambda)
    \mathbf{D}_{\boldsymbol{\alpha}}^{-1}.
\end{align}
\end{proof}

\section{KFAC Congruence under ReLU Rescaling}
\label{proof:kfac}

We verify that standard KFAC curvature blocks obey the same congruence law as
Jacobian-based curvature under the ReLU rescalings of Section~\ref{sec:relu-orbits}. Consider an affine layer
\[
\mathbf{z}_{\ell}
=
\mathbf{W}_{\ell}\mathbf{a}_{\ell-1}
+
\mathbf{b}_{\ell},
\qquad
\mathbf{a}_{\ell}
=
\phi(\mathbf{z}_{\ell}).
\]
Let
\[
\mathbf{D}_{\ell}
=
\operatorname{diag}(\alpha_{\ell,1},\ldots,\alpha_{\ell,d_{\ell}})
\]
collect the positive hidden-unit scaling factors at layer \(\ell\), with
\(\mathbf{D}_{0}=\mathbf{D}_{L}=\mathbf{I}\) because input and output units are not rescaled.
The ReLU parameter rescaling acts layerwise as
\[
\mathbf{W}_{\ell}'
=
\mathbf{D}_{\ell}\mathbf{W}_{\ell}\mathbf{D}_{\ell-1}^{-1},
\qquad
\mathbf{b}_{\ell}'
=
\mathbf{D}_{\ell}\mathbf{b}_{\ell}.
\]
This is the layerwise matrix form of the elementary hidden-unit rescaling in Eq.~\ref{eq:elementary-rescaling}. With
column-major vectorization,
\[
\operatorname{vec}(\mathbf{W}_{\ell}')
=
\mathbf{S}_{\ell}\operatorname{vec}(\mathbf{W}_{\ell}),
\qquad
\mathbf{S}_{\ell}
=
\mathbf{D}_{\ell-1}^{-1}\otimes \mathbf{D}_{\ell},
\]
so \(\mathbf{S}_{\ell}\) is the corresponding parameter-space block of the global rescaling matrix
\(\mathbf{D}_{\boldsymbol{\alpha}}\).

The parameter rescaling induces the activation transformation
\[
\mathbf{z}_{\ell}'
=
\mathbf{D}_{\ell}\mathbf{z}_{\ell},
\qquad
\mathbf{a}_{\ell}'
=
\mathbf{D}_{\ell}\mathbf{a}_{\ell},
\]
where the second identity follows from positive homogeneity,
\(\phi(\mathbf{D}_{\ell}\mathbf{z}_{\ell})=\mathbf{D}_{\ell}\phi(\mathbf{z}_{\ell})\).
For the loss derivative
\[
\boldsymbol{\delta}_{\ell}
=
\partial \mathcal{L}/\partial \mathbf{z}_{\ell},
\]
the chain rule gives
\[
\boldsymbol{\delta}_{\ell}'
=
\mathbf{D}_{\ell}^{-1}\boldsymbol{\delta}_{\ell}.
\]

With column-major vectorization, the KFAC approximation to the weight block is
\[
\mathbf{C}_{\ell}^{\mathrm{KFAC}}
=
\mathbf{A}_{\ell-1}\otimes\mathbf{G}_{\ell},
\qquad
\mathbf{A}_{\ell-1}
=
\mathbb{E}\!\left[
\mathbf{a}_{\ell-1}\mathbf{a}_{\ell-1}^{\top}
\right],
\qquad
\mathbf{G}_{\ell}
=
\mathbb{E}\!\left[
\boldsymbol{\delta}_{\ell}\boldsymbol{\delta}_{\ell}^{\top}
\right].
\]
The transformations above imply
\[
\mathbf{A}_{\ell-1}'
=
\mathbf{D}_{\ell-1}\mathbf{A}_{\ell-1}\mathbf{D}_{\ell-1},
\qquad
\mathbf{G}_{\ell}'
=
\mathbf{D}_{\ell}^{-1}\mathbf{G}_{\ell}\mathbf{D}_{\ell}^{-1}.
\]
Therefore, by the mixed-product property of the Kronecker product,
\[
\mathbf{C}_{\ell}^{\mathrm{KFAC}\prime}
=
\mathbf{A}_{\ell-1}'\otimes\mathbf{G}_{\ell}'
=
(\mathbf{D}_{\ell-1}\otimes\mathbf{D}_{\ell}^{-1})
\mathbf{C}_{\ell}^{\mathrm{KFAC}}
(\mathbf{D}_{\ell-1}\otimes\mathbf{D}_{\ell}^{-1})
=
\mathbf{S}_{\ell}^{-1}
\mathbf{C}_{\ell}^{\mathrm{KFAC}}
\mathbf{S}_{\ell}^{-1}.
\]
This is the layerwise form of the global congruence law
\[
\mathbf{C}_{v}
=
\mathbf{D}_{\boldsymbol{\alpha}}^{-1}
\mathbf{C}_{w}
\mathbf{D}_{\boldsymbol{\alpha}}^{-1}.
\]

The KFAC factorization is compatible with the rescaling because the factorized expectation commutes
with the induced activation and derivative transformations:
\[
\mathbb{E}\!\left[
\mathbf{a}_{\ell-1}'\mathbf{a}_{\ell-1}'^{\top}
\right]
\otimes
\mathbb{E}\!\left[
\boldsymbol{\delta}_{\ell}'\boldsymbol{\delta}_{\ell}'^{\top}
\right]
=
(\mathbf{D}_{\ell-1}\otimes\mathbf{D}_{\ell}^{-1})
\left(
\mathbb{E}\!\left[
\mathbf{a}_{\ell-1}\mathbf{a}_{\ell-1}^{\top}
\right]
\otimes
\mathbb{E}\!\left[
\boldsymbol{\delta}_{\ell}\boldsymbol{\delta}_{\ell}^{\top}
\right]
\right)
(\mathbf{D}_{\ell-1}\otimes\mathbf{D}_{\ell}^{-1}).
\]

Biases are handled by augmenting the activation vector,
\[
\bar{\mathbf{a}}_{\ell-1}
=
(\mathbf{a}_{\ell-1}^{\top},1)^{\top},
\qquad
\bar{\mathbf{D}}_{\ell-1}
=
\operatorname{diag}(\mathbf{D}_{\ell-1},1).
\]
The appended constant coordinate is left unscaled because it is not a hidden unit in the ReLU scale
group, while the bias column itself scales with \(\mathbf{D}_{\ell}\). Equivalently, for the augmented
weight matrix \(\bar{\mathbf{W}}_{\ell}=[\mathbf{W}_{\ell}\ \mathbf{b}_{\ell}]\),
\[
\bar{\mathbf{W}}_{\ell}'
=
\mathbf{D}_{\ell}\bar{\mathbf{W}}_{\ell}\bar{\mathbf{D}}_{\ell-1}^{-1}.
\]
The same congruence argument then applies to the augmented weight-bias block.

For convolutional layers in the standard unfolded-patch KFAC formulation, the proof carries over
after replacing \(\mathbf{a}_{\ell-1}\) by the unfolded input patches and applying the corresponding
channel scaling to each spatial patch coordinate.
\section{Experimental Details}
\label{app:exp_details}
Our experiments are fully reproducible. All the experiments are conducted on a single NVIDIA A30 24GB GPU. Code is available in the supplementary material.
\subsection{Orbit Stress Test}
\paragraph{Dataset and model}
We use Fashion-MNIST~\cite{xiao2017fashion}, with $60{,}000$ training images and $10{,}000$
test images across 10 classes. Images are normalized using the training-set mean $0.2860$ and
standard deviation $0.3530$, and are flattened to vectors of dimension $784$.
The model is a two-hidden-layer ReLU MLP with hidden dimensions $[64, 64]$ and a linear
output layer. It has no normalization layers. 

\paragraph{MAP training}
The MAP solution is obtained by minimizing cross-entropy with $\ell_2$ weight decay $10^{-4}$
using Adam with learning rate $10^{-3}$, batch size $256$, for $10$ epochs. The resulting
MAP model achieves $87.67\%$ test accuracy.

\paragraph{Laplace approximation}
We use \texttt{laplace-torch}~\cite{daxberger2021laplace}. All Laplace fits use all
parameters, diagonal GGN curvature, and a scalar prior precision $\lambda$. The prior
precision is initialized at $1.0$ and optimized by $100$ Adam steps with learning rate $0.1$
on the Laplace marginal likelihood. For prediction, we use the Jacobian-linearized predictive
distribution with the probit link approximation. Model computations use \texttt{float32};
$\gamma$-canonicalization is performed in \texttt{float64} and cast back to \texttt{float32}
before fitting Laplace.

\paragraph{Orbit generation}
Starting from the MAP solution $\hat{w}$, we generate $K=20$ function-preserving
representatives. For each representative, we sample independent hidden-unit log-scales
\[
  \log\alpha_i \overset{\mathrm{i.i.d.}}{\sim} \mathcal{N}(0,\,\sigma^2),\quad \sigma = 1.0,
\]
and apply the elementary ReLU rescaling in Eq.~\ref{eq:elementary-rescaling} to every hidden unit. For every generated
representative $v_k$, we verify
\[
  \max_{x\in\mathcal{D}_{\mathrm{test}}} \bigl\|f(x;\hat{w}) - f(x;v_k)\bigr\|_\infty < 10^{-5}.
\]
Thus all reported variation across orbit samples is due to the parameterization supplied to the
Laplace approximation, not to any change in the predictor.

\paragraph{$\gamma$-canonicalization}
For each representative \(v_k\) and each
\(\gamma \in \{0,\,0.25,\,0.50,\,0.75,\,1.0\}\), we apply Algorithm~1 with tolerance
\(\varepsilon=10^{-6}\) and a maximum budget of \(200\) sweeps. For interior values, all runs converged in fewer than \(50\) sweeps under the stopping
criterion \(\max_{\ell,i}|\log \alpha_{\ell,i}|<10^{-6}\). The endpoints are computed by a single
directional pass: forward incoming-normalization for \(\gamma=0\) and backward outgoing-normalization
for \(\gamma=1\). The canonicalization cost is much smaller than the
cost of fitting the Laplace approximation.

\paragraph{Metrics}
We report the following quantities. NLL is the mean negative log-likelihood of the Laplace
predictive distribution on the test set. ECE is the expected calibration error with 20
equal-width confidence bins~\cite{guo2017calibration}. The quantity $\log Z$ is the Laplace
marginal likelihood at the selected $\lambda^\star$. The norm $\|w\|^2$ is the squared
Euclidean norm of the parameter vector supplied to Laplace; for standard EB this is the raw
(possibly rescaled) representative, and for $\gamma$-canonicalized EB it is the canonical
representative. The statistic $\max\|\Delta p\|_\infty$ is the maximum absolute difference
in predicted class probability between any two orbit samples, computed over all $\binom{K}{2}$
pairs, all test inputs, and all classes. The spread factor is
$\exp\!\bigl(\max_k \log\lambda^\star_k - \min_k \log\lambda^\star_k\bigr)$.

\subsection{CIFAR-10 ConvNet with KFAC-Laplace}
\label{app:cifar10-kfac}

\paragraph{Model and training.}
We use a plain ReLU ConvNet without normalization layers. The channel widths are $(64,128,256,256)$. The network consists of
four convolutional blocks followed by global average pooling and a linear
classifier. Each convolutional block contains a $3\times 3$ convolution followed
by a ReLU nonlinearity; spatial downsampling is performed by avg pooling. The
model is trained on CIFAR-10 using Adam with learning rate $10^{-3}$ and weight
decay $10^{-5}$ for $50$ epochs. We use a cosine learning-rate schedule.

\paragraph{Laplace approximation.}
We fit all-layer empirical-Bayes Laplace with KFAC curvature using
\texttt{laplace-torch}. The prior precision is optimized by marginal likelihood
using \texttt{method="marglik"}. Predictions use the Jacobian-linearized
predictive distribution with the probit link approximation. For the
canonicalized variant, we first apply $\gamma$-canonicalization to the trained
MAP solution and then fit the same KFAC-Laplace approximation. We evaluate
$
    \gamma \in \{0,0.1,0.2,\ldots,1.0\}
$
and select $\gamma$ by marginal likelihood within the canonicalized family. In
all three seeds, the selected value is $\gamma=0.5$.

\section{Within-orbit vs. between-solution variation}
We compare variation between $20$ function-preserving rescalings
of one trained network ($\sigma=1$), and $20$ independently trained
networks with the same architecture and training procedure.
We fit standard EB-Laplace to every rescaled or independently trained
network. Table~\ref{tab:rescaling_retraining} shows that the standard
deviation of $\log\lambda^\star$ is approximately $22$ times larger
under rescaling than retraining. For $\log Z$, it is approximately
$10$ times larger.
Predictive metrics also vary under rescaling, although retraining
produces larger variation in NLL and accuracy.
Variation across training runs can reflect different learned functions.
In contrast, variation under function-preserving rescaling arises
entirely from the parameter scales supplied to the Laplace fit.
These results show that arbitrary scale choices can affect the selected
prior precision and evidence more than independent retraining.

\begin{table}[t]
\centering
\caption{
Standard deviations under rescaling and retraining.
Rescaling preserves the network function while retraining produces
different learned functions. Predictive metrics use standard EB-Laplace.
}
\label{tab:rescaling_retraining}
\small
\setlength{\tabcolsep}{8pt}
\begin{tabular}{lrr}
\toprule
Quantity & Rescaling & Retraining \\
\midrule
Log prior precision $\log\lambda^\star$
& $0.356$ & $0.016$ \\
Log evidence $\log Z$
& $8900$ & $884$ \\
NLL
& $0.0045$ & $0.0088$ \\
ECE
& $0.0059$ & $0.0046$ \\
Accuracy
& $0.0008$ & $0.0040$ \\
Squared parameter norm $\|w\|_2^2$
& $1016.0$ & $4.7$ \\
\bottomrule
\end{tabular}
\end{table}

\end{document}